\documentclass[10pt,twocolumn,letterpaper]{article}

\usepackage{cvpr}              

\usepackage{graphicx}

\usepackage{booktabs}

\usepackage{multirow}

\usepackage{microtype}

\usepackage{siunitx}

\usepackage{caption}

\usepackage{wrapfig}

\usepackage{ragged2e}

\usepackage{url}

\usepackage{colortbl}

\usepackage{amsmath,amsfonts,bm}

\def\eqref#1{equation~\ref{#1}}

\def\1{\bm{1}}

\DeclareMathAlphabet{\mathsfit}{\encodingdefault}{\sfdefault}{m}{sl}
\SetMathAlphabet{\mathsfit}{bold}{\encodingdefault}{\sfdefault}{bx}{n}

\definecolor{cvprblue}{rgb}{0.21,0.49,0.74}
\usepackage[pagebackref,breaklinks,colorlinks,allcolors=cvprblue]{hyperref}

\def\paperID{7369} 
\def\confName{CVPR}
\def\confYear{2026}

\title{HyperST: Hierarchical Hyperbolic Learning for

Spatial Transcriptomics Prediction}

\author{
Chen Zhang$^{1*}$ \quad 
Yilu An$^{1*}$ \quad 
Ying Chen$^{1,2*}$ \quad
Hao Li$^{1}$ \quad
Xitong Ling$^{3}$ \quad
Lihao Liu$^{2}$ \\
Junjun He$^{2}$ \quad
Yuxiang Lin$^{1}$\footnotemark[2] \quad
Zihui Wang$^{4}$\footnotemark[2] \quad
Rongshan Yu$^{1}\footnotemark[2]$\\ \\
$^{1}$Xiamen University \quad 
$^{2}$Shanghai AI Laboratory  \quad
$^{3}$Tsinghua University \quad
$^{4}$Peng Cheng Laboratory \quad
}

\begin{document}
\maketitle

\let\thefootnote\relax\footnotetext{*These authors contributed equally to this work.}
\let\thefootnote\relax\footnotetext{\dag Corresponding authors: Yuxiang Lin (linyuxiang@stu.xmu.edu.cn), Zihui Wang  (wangzh08@pcl.ac.cn), Rongshan Yu (rsyu@xmu.edu.cn)}

\begin{abstract}
Spatial Transcriptomics (ST) merges the benefits of pathology images and gene expression, linking molecular profiles with tissue structure to analyze spot-level function comprehensively.
Predicting gene expression from histology images is a cost-effective alternative to expensive ST technologies.
However, existing methods mainly focus on spot-level image-to-gene matching but fail to leverage the full hierarchical structure of ST data, especially on the gene expression side, leading to incomplete image-gene alignment.
Moreover, a challenge arises from the inherent information asymmetry: gene expression profiles contain more molecular details that may lack salient visual correlates in histological images, demanding a sophisticated representation learning approach to bridge this modality gap.
We propose \textbf{HyperST}, a framework for ST prediction that learns multi-level image-gene representations by modeling the data's inherent hierarchy within hyperbolic space, a natural geometric setting for such structures.
First, we design a \textbf{Multi-Level Representation Extractors} to capture both spot-level and niche-level representations from each modality, providing context-aware information beyond individual spot-level image-gene pairs.
Second, a \textbf{Hierarchical Hyperbolic Alignment} module is introduced to unify these representations, performing spatial alignment while hierarchically structuring image and gene embeddings.
This alignment strategy enriches the image representations with molecular semantics, significantly improving cross-modal prediction.
HyperST achieves state-of-the-art performance on four public datasets from different tissues, paving the way for more scalable and accurate spatial transcriptomics prediction. 

\end{abstract}

\begin{figure}[htbp] 
    \centering
 
    \includegraphics[width=1\columnwidth]{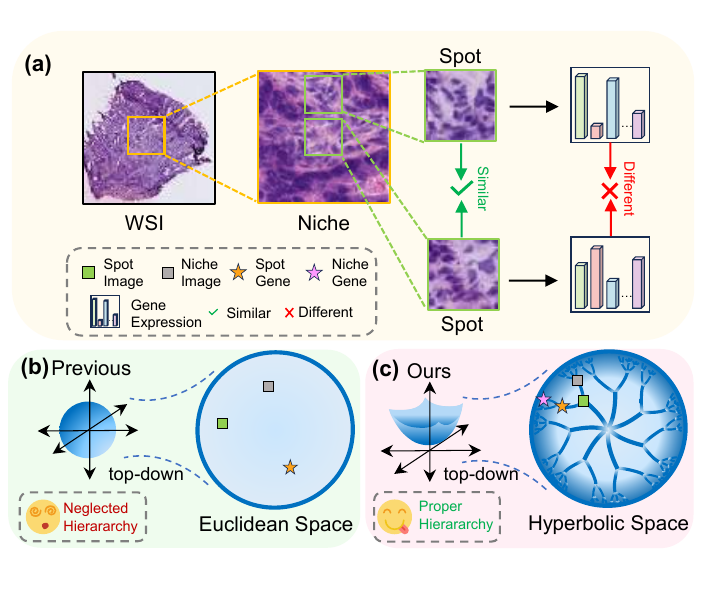}

    \caption{\textbf{ST data characteristics}. (a) A WSI 
    contains hierarchical structures and visually similar 
    patterns may correspond to different gene expression profiles. 
    (b) Other works mainly model ST data in Euclidean Space, 
    which neglects niche-level gene and can lead to 
    biased biological insights. (c) Our hyperbolic approach 
    models hierarchies based on information specificity, 
    where a general concept (image/spot) entails its more 
    specific, information-rich counterpart (gene/niche), 
    enabling more informative representation learning.}
    \label{fig1}
\end{figure}

\section{Introduction}\label{intro}

Pathological images, particularly Hematoxylin and Eosin (H\&E) stained Whole Slide Images (WSIs), provide critical insights into cell morphology and tissue architecture, serving as a cornerstone in biomedical research and clinical diagnosis~\cite{chen2025slidechat, lu2024visual,chen2024towards,li2024generalizable}.
Gene expression data complement these pathological images by elucidating the molecular mechanisms underlying observed features, thereby enhancing disease diagnosis and facilitating therapeutic target identification~\cite{ash2021joint}.
Spatial Transcriptomics (ST) integrates both modalities by capturing spatially resolved gene expression and cellular morphology simultaneously~\cite{staahl2016visualization}, aligning molecular profiles with tissue structure at micrometer resolution~\cite{williams2022introduction}.
Despite its advantages, ST has not achieved widespread clinical adoption due to its high cost and laborious experiment compared to traditional techniques~\cite{zhang2022clinical, choe2023advances}. Consequently, there has been increasing attention on predicting spatially resolved gene expression directly from pathological images using deep learning approaches~\cite{wang2025benchmarking}.

Recent studies have explored diverse strategies for this prediction task, including direct inference from spot-level images~\cite{he2020integrating, monjo2022efficient}, integration of multi-scale features across WSIs~\cite{chung2024accurate, wang2025m2ost}, and contrastive learning to align spot-level images with gene expression profiles~\cite{xie2023spatially}. Although these methods have shown promising results, several critical questions remain underexplored.
Our study is motivated by two key questions in ST prediction: 
(1) \textit{\textbf{Can integrating broader pathological and genetic context improve spot-level gene expression inference?}} 
Previous studies often  primarily utilized multi-scale pathological features for gene expression prediction, neglecting the multi-level structure inherent in gene expression itself~\cite{jaume2024modeling,chen2024survmamba}, which spans cellular and tissue-level scales. In reality, both broader pathological context and bulk genetic programs can significantly influence the gene expression profile at each spot~\cite{chen2020spatial, nirmal2022spatial,wu2025metsdb, ye2024single}.
(2) \textit{
\textbf{How to encode more molecular information from images under the discrepancy between visual similarity and molecular heterogeneity?}
} As illustrated in Figure \ref{fig1} (a), biological heterogeneity may result in visually similar pathology patches exhibiting distinct gene expression patterns~\cite{zhu2025diffusion,pizurica2024digital,fujii2024decoding,tang2025spatial}. 
This phenomenon indicates that standard image encoders may fail to capture the subtle morphological cues for predicting these molecular variations. Rather than falling into a one-to-many formulation, we focus on learning a more powerful and molecularly-informed image representation to better capture these complex relationships.

To address these two questions, we introduce \textbf{HyperST}, a novel framework for ST prediction by learning multi-level hyperbolic image-gene representations.
HyperST tackles these challenges with two core components.
First, our \textbf{Multi-Level Representation Extractors} capture hierarchical representations from both pathology images and their corresponding gene expression profiles. They extract multimodal information at both spot level and niche level, where a niche consists of a central spot and its surrounding neighbors, enabling the capture of comprehensive morphological and molecular patterns across spatial scales.
Second, our \textbf{Hierarchical Hyperbolic Alignment} module acts as a powerful structural regularizer rather than a generative model. It uses the unique properties of hyperbolic geometry to impose a meaningful inductive bias on the latent space, guiding the model to learn molecularly-informed features.
%


We define our hierarchical relationships based on \textbf{information specificity}.
In this view, a concept A entails a concept B if B is a semantically richer and more specific instance of A. For example, the concept of a ``dog on a beach'' is more specific and information-rich than ``dog'', and is thus considered the child concept. This viewpoint has been adopted in hyperbolic representation learning~\cite{desai2023hyperbolic, ge2023hyperbolic}.
Following this principle, we establish two key hierarchies in our framework: 
(1) A spot-level representation entails its context-rich niche-level counterpart.
(2) A morphological image entails its corresponding gene expression. This is because the gene profile contains fine-grained molecular information that offers a much more specific description of the tissue's state than the more general pathology image.
HyperST learns powerful, context-aware representations by modeling information-based hierarchies in hyperbolic space, which is inherently more suited for capturing such structures than Euclidean space~\cite{hsu2021capturing}.

We demonstrate HyperST's effectiveness on four public datasets from diverse tissues, where it consistently outperforms state-of-the-art models. Our contributions are summarized as:


%
\begin{itemize}
    \item We propose \textbf{HyperST}, a novel framework for predicting spatially resolved gene expression from WSIs by learning multi-level hyperbolic representations that capture the intrinsic hierarchical structure of ST data.
    \item We design \textbf{Multi-Level Representation Extractors} to capture spot- and niche-level representations from both modalities, providing comprehensive biological insights.
    \item We introduce \textbf{Hierarchical Hyperbolic Alignment} to structurally regularize the latent space, improving cross-modal feature integration.

    \item Extensive experiments on four public datasets demonstrate that HyperST consistently outperforms existing approaches, underscoring its robust efficacy in spatial gene expression prediction.
\end{itemize}

\section{Related Work}

\subsection{Prediction of gene expression from images}

Recent methodologies for predicting spatially resolved gene expression from histology images have advanced through diverse computational paradigms, including StNet~\citep{he2020integrating}, BLEEP~\citep{xie2023spatially}, TRIPLEX~\citep{chung2024accurate}, and Stem~\citep{zhu2025diffusion}. Local image-to-expression regression models like StNet employ ResNet50~\citep{he2016deep} to directly map H\&E image patches to gene expression profiles. While effective in deterministic prediction, these methods assume injective mappings between morphology and transcription, overlooking biological heterogeneity. Multi-scale integration approaches like TRIPLEX extract and fuse multi-resolution features from WSIs using attention mechanisms. Although these methods capture multi-resolution visual patterns, they lack explicit constraints to preserve the essential biological hierarchy. Generative models like Stem address the uncertainty in expression prediction by generating probabilistic gene expression profiles. While these paradigms better preserve transcriptional variability, they neglect the inherent data hierarchy. In contrast to these prior works, HyperST explicitly models the intrinsic parent-child relationships between spots and their surrounding niches across both imaging and gene expression modalities.

\subsection{Multimodal representation learning}

Contrastive learning is a pivotal technique for cross-modal tasks by aligning representations across different modalities. For example, CLIP~\citep{radford2021learning} employs contrastive learning to align paired images and texts in a shared Euclidean embedding space. Inspired by CLIP, BLEEP~\citep{xie2023spatially} adapts contrastive learning to histology and gene expression, using direct interpolation in the embedding space for efficient, decoder-free predictions. These two models rely on Euclidean embeddings, which limit their ability to capture hierarchical relationships. To overcome these limitations, MERU~\citep{desai2023hyperbolic} embeds image and text into hyperbolic space, leveraging its geometric properties to build a hierarchical representation space through contrastive and entailment losses. Building on MERU, HyCoCLIP~\citep{pal2024compositional} introduces intra-modal hierarchical modeling by extracting object bounding boxes from images and their corresponding textual descriptions, establishing hierarchical links between box regions and the full image-text pair. HyCoCLIP’s dependence on pre-trained object detection models to derive these boxes from given captions may result in potential inaccuracies. In contrast, HyperST directly leverages the inherent structure of ST data, from spot-level to niche-level contexts, avoiding uncertainties associated with external feature extraction. 
\section{Method}\label{meth}

\begin{figure*}[t] 
    \centering
    \includegraphics[width=1\textwidth]{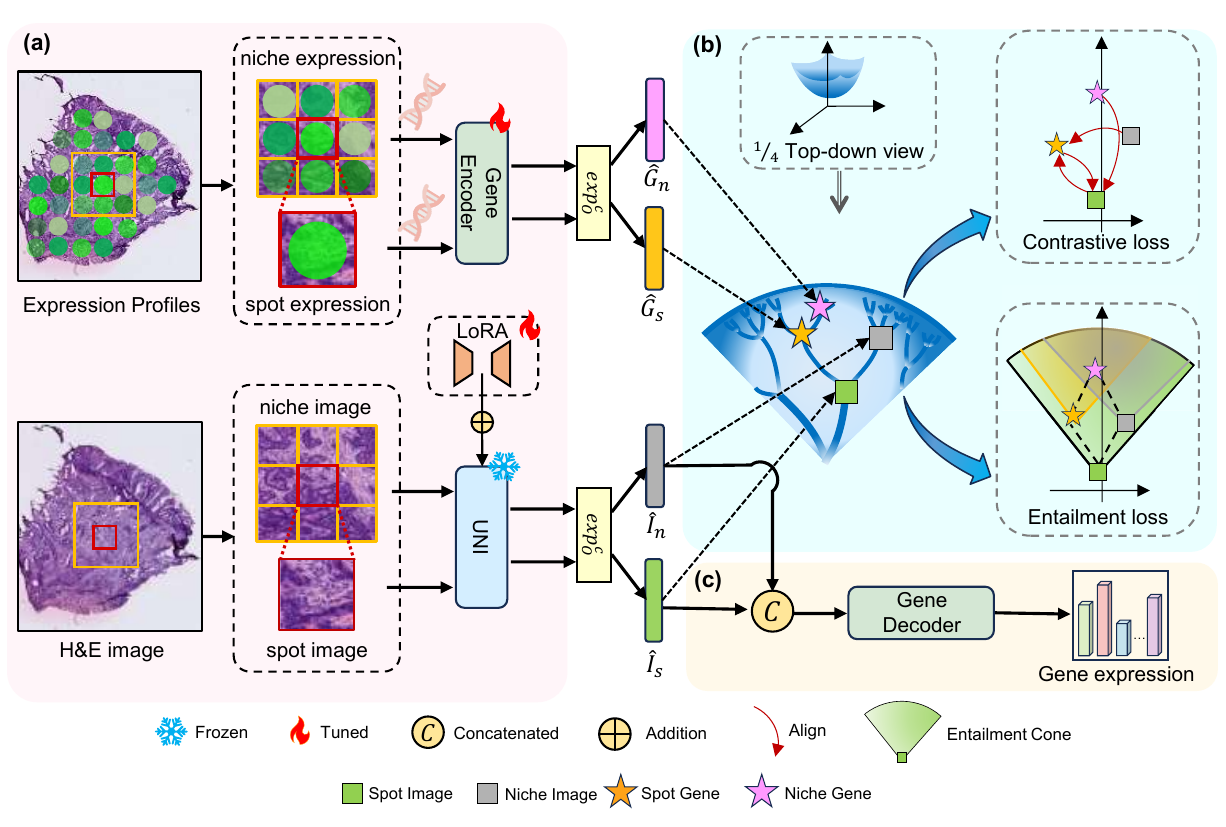}  
    \caption{\textbf{Overview of HyperST.} HyperST consists of three components. (a) \textbf{Multi-Level Representation Extractors} capture spot- and niche-level features from both images and gene expression. (b) \textbf{Hierarchical Hyperbolic Alignment} module projects these features into a shared hyperbolic latent space. It uses contrastive alignment for corresponding image-gene pairs and entailment alignment to structurally regularize the latent space according to information hierarchies. (c) \textbf{Gene Decoder} uses the resulting aligned and context-aware image representations to predict spot-level gene expression.}  
    \label{fig2}
\end{figure*}

\vspace{-2mm}

The overview of HyperST is illustrated in Figure \ref{fig2}. First, we briefly describe the preliminaries of the hyperbolic geometry in Section \ref{Section 3.1}. Second, we present the Multi-Level Representation Extractors in Section \ref{Section 3.2}. Third, we introduce the Hierarchical Hyperbolic Alignment in Section \ref{Section 3.3}. Finally, we describe the Gene Decoder and our Overall Objective Function in Section \ref{Section 3.4}.

\subsection{Preliminaries}
\label{Section 3.1}

\paragraph{Hyperbolic Geometry} Hyperbolic Geometry is a fundamental class of non-Euclidean geometry with a constant negative curvature. This distinguishing characteristic results in an exponential growth of volume with respect to radius, in stark contrast to Euclidean geometry, which exhibits zero curvature and polynomial volume scaling~\citep{mettes2024hyperbolic}. Consequently, hyperbolic spaces are naturally adept at representing tree-like or hierarchical data structures, where the number of elements increases exponentially with depth~\citep{hsu2021capturing, pal2024compositional}. Due to their negative curvature, hyperbolic spaces cannot be isometrically embedded in Euclidean spaces of equivalent dimensionality without compromising distances or angles. To address this issue, several geometric models are employed for their representation and computation, including the Poincaré ball model and the Lorentz model~\citep{cannon1997hyperbolic,cho2022rotated}. 


\paragraph{Lorentz Model} Lorentz model is widely preferred due to its numerical stability and straightforward geodesic calculations~\citep{nickel2018learning}. The Lorentz model $\mathbb{L}^n_c$ embeds the $n$-dimensional hyperbolic space as the upper sheet of a two-sheeted hyperboloid in $(n+1)$-dimensional Minkowski space, with a constant curvature $-c < 0$. It therefore consists of all vectors satisfying the following conditions:

\begin{equation}
  \label{hyperbolic denfine} 
  \begin{aligned}
    \mathbb{L}^n_c = \{ \mathbf{x} \in \mathbb{R}^{n + 1} : & \langle \mathbf{x}, \mathbf{x} \rangle_{\mathbb{L}}\ = - \frac{1}{c} , \\
                                         & x_{time}=\sqrt{1/c+\| \mathbf{x}_{space}\|^2}, \\
                                         & c > 0 \} , 
  \end{aligned}
\end{equation}
where points $\mathbf{x}\in \mathbb{R}^{n + 1}$ in $\mathbb{L}^n_c$ can be represented as $[x_{time}, \mathbf{x}_{space}]$. $x_{time} \in \mathbb{R}$ and $\mathbf{x}_{space}\in \mathbb{R}^{n}$ denote the \textit{time component} and the \textit{spatial component}~\citep{desai2023hyperbolic}, respectively. For two vectors $\mathbf{x}, \mathbf{y} \in \mathbb{L}^{n}_c$, the Lorentzian inner product $\langle \cdot, \cdot \rangle_{\mathbb{L}}$ is defined as $\langle \mathbf{x}, \mathbf{y} \rangle_{\mathbb{L}} = \langle \mathbf{x}_{space}, \mathbf{y}_{space}\ \rangle_{\mathbb{E}} -x_{time}y_{time}$ ,
where $\langle \mathbf{x}, \mathbf{y} \rangle_{\mathbb{E}}$ represents the Euclidean inner product in $\mathbb{R}^n$. Besides, the Lorentzian distance $d_{\mathbb{L}}(\mathbf{x}, \mathbf{y})$ measures the length of the shortest path between two points $\mathbf{x}$ and $\mathbf{y}$, which is formulated as:
\begin{equation}
    d_{\mathbb{L}}(\mathbf{x}, \mathbf{y}) = \sqrt{1/c} \cdot \cosh^{-1}(-c\langle \mathbf{x}, \mathbf{y} \rangle_{\mathbb{L}}).
\label{distence of hyperbolic}
\end{equation} 

\paragraph{Tangent Space and Exponential Map} The tangent space of $ \mathbf{x} \in \mathbb{L}^n$ is denoted by $\mathcal{T}_{\mathbf{x}}\mathbb{L}^n_c$, which is precisely defined as the set of vectors orthogonal to $\mathbf{x}$ under the Lorentzian inner product:
\begin{equation}
    \mathcal{T}_{\mathbf{x}}\mathbb{L}^n_c=\{\mathbf{v} \in \mathbb{R}^{n+1}:\langle 
    \mathbf{x}, \mathbf{v}\rangle_{\mathbf{L}} = 0\}.
\end{equation} 
A fundamental mechanism for connecting the tangent space to the hyperbolic manifold is the exponential map. 
The exponential map $\exp^c_\mathbf{x}:\mathcal{T}_{\mathbf{x}}\mathbb{L}^n_c \to \mathbb{L}^n_c$ projects tangent vector $\mathbf{v} $ onto the $\mathbb{L}^n_c$ along a geodesic emanating from $\mathbf{x}$ in the direction of $\mathbf{v}$, given by: 
\begin{equation}
    \exp^c_\mathbf{x}(\mathbf{v})=\cosh (\sqrt{c}\|\mathbf{v}\|_{\mathbb{L}})\mathbf{x} + \frac{\sinh(\sqrt{c}\|\mathbf{v}\|_{\mathbb{L}})}{\sqrt{c}\|\mathbf{v}\|_{\mathbb{L}}}\mathbf{v},
\end{equation} 
where $\|\mathbf{v}\|_{\mathbb{L}}=\sqrt{\langle \mathbf{v}, \mathbf{v} \rangle}_{\mathbb{L}}$ is the Lorentzian norm. Moreover, the exponential map serves as a bridge between Euclidean and hyperbolic geometries. By interpreting Euclidean vectors as tangent vectors at the origin $\mathbf{O}=[\sqrt{1/c}, 0, \dots, 0] \in \mathbb{R}^{n + 1}$ of the hyperbolic space~\citep{mettes2024hyperbolic, pal2024compositional, khrulkov2020hyperbolic}, we begin by extending the Euclidean embedding $\mathbf{v}_{euc} \in \mathbb{R}^n$ into $\mathbb{R}^{n + 1}$ by defining a vector $\mathbf{v}=[0, \mathbf{v}_{euc}] \in \mathbb{R}^{n + 1 }$. This vector $\mathbf{v}$ is situated in the tangent space at the origin $\mathbf{O}$ of the hyperboloid as $\langle \mathbf{O}, \mathbf{v} \rangle_{\mathbb{L}}=0$. Thus, $\mathbf{v}$ can be projected onto the hyperboloid $\mathbb{L}^n_c$ employing the exponential map:
\vspace{0.2mm}
\begin{equation}
    \mathbf{x}_{space} = \exp_{\mathbf{O}}^{c}(\mathbf{v}_{euc}) = \frac{\sinh(\sqrt{c} \|\mathbf{v}_{euc}\|_{\mathbb{E}})}{\sqrt{c}\|\mathbf{v}_{euc}\|_{\mathbb{E}}}\mathbf{v}_{euc}.
\label{exponential map}
\end{equation}
Then we can directly calculate the corresponding time component $x_{time}$  from $\mathbf{x}_{space}$. The detailed derivations of the above equations can be found in Section 6 of the Supplementary Material.

\paragraph{Hyperbolic Entailment Loss} The entailment cone $\mathcal{R}_\mathbf{y}$ constitutes a region around the point $\mathbf{y}$ where all points $\mathbf{x} \in \mathcal{R}_{\mathbf{y}}$ represent child concepts of the parent concept $\mathbf{y}$ ~\citep{ganea2018hyperbolic,desai2023hyperbolic}, defined by the half-aperture:
\begin{equation}
    \text{aper}(\mathbf{y}) = \sin^{-1}\Bigg(\frac{2K}{\sqrt{c} \| \mathbf{y}_{space} \|}\Bigg),
\end{equation}
where $K=0.1$ determines boundary conditions near the origin. To enforce the partial order relationship where $\mathbf{y}$ entails $\mathbf{x}$, the penalty is formulated as:
\begin{equation}
    \mathcal{L}_{entail}(\mathbf{y}, \mathbf{x}) = \max{(0, \text{ext}(\mathbf{y}, \mathbf{x})- \text{aper}(\mathbf{y}))},
\label{entail loss}
\end{equation}
where $\text{ext}(\mathbf{y}, \mathbf{x})$ denotes the exterior angle defined as $\text{ext}(\mathbf{y}, \mathbf{x}) = \cos^{-1}\Big(\frac{x_{time} + y_{time}c\langle \mathbf{y}, \mathbf{x} \rangle_{\mathbb{L}}}  {\|\mathbf{y}_{space} \| \sqrt{(c \langle \mathbf{y}, \mathbf{x} \rangle_{\mathbb{L}})^2-1}}\Big)$.

\subsection{Multi-Level Representation Extractors}
\label{Section 3.2}

\paragraph{Multi-Level Pathological Images Extractor}
Following the previous works~\citep{xie2023spatially, zhu2025diffusion}, a spot-level image patch $X_{s} \in \mathbb{R}^{3 \times L_s \times L_s}$ of each identified spot is extracted and preprocessed from a H$\&$E stained image, with the spot positioned at the center of the patch, as depicted in Figure \ref{fig2} (a), where $L_s$ represents the size of the spot image. While $X_{s}$ directly corresponds to the target spot gene expression, the additional nearby visual information from larger-scale pathology patches can significantly contribute to the analysis~\citep{chung2024accurate, lin2024st}. Therefore, we introduce the niche-level image patch $X_{n} \in \mathbb{R}^{3 \times L_n \times L_n}$, which is defined as a higher-level region composed of the central spot-level patch $X_{s}$ and its spatially adjacent spot-level patches. $L_n$ signifies the patch size of the niche image. These neighboring patches are selected based on spatial proximity using the K-Nearest Neighbors (KNN) algorithm. By cropping the region of these patches, $X_{n}$ forms a larger image region that provides a broader field of view and enhanced contextual information about the surrounding tissue microenvironment.

We leverage UNI~\citep{chen2024towards}, a pathology foundation model pre-trained on large-scale histology images, to extract feature embeddings for spot-level and niche-level image patches.
As the original UNI was not well-suited for large-sized niche-level image patches, we resized the images in our dataset and fine-tuned UNI using the Low-Rank Adaptation (LoRA) technique~\citep{hu2022lora}, leading to improved multi-level visual representations. Consider a frozen pre-trained weight matrix $W_{origin}\in \mathbb{R}^{d \times d}$, where $d$ denotes the dimension. The updated weight matrix is formulated as $W_{new} = W_{origin} + \Delta W = W_{origin} + BA$,
where the update $\Delta W \in \mathbb{R}^{d \times d}$ expressed by the product of two smaller trainable matrices: $B\in \mathbb{R}^{d \times r}$, $A\in \mathbb{R}^{r \times d}$, and the rank $r \ll d $. This approach enables us to adapt the model to the characteristics of our data while substantially reducing the computational resources required for fine-tuning. The multi-level image representations $I_{s} \in \mathbb{R}^{d}$ and $I_{n} \in \mathbb{R}^{d}$ are extracted by $I_{s}, I_{n}=MIE(X_{s}, X_{n} ;\theta_{UNI}, \Delta\theta_{lora})$.
Here, $MIE$ represents the UNI model adapted by LoRA, $\theta_{uni}$ and $\Delta \theta_{lora}$ denote the frozen parameters of UNI and the trainable parameters of LoRA modules.

\begin{table*}[t] 
\centering
\footnotesize
\caption{Performance comparison on four spatial transcriptomics datasets. Higher values on PCC@10, PCC@50, PCC@200 are better. Lower values on MAE, MSE are better.}
\label{tab:baseline}

\newcommand{\std}[1]{{\scriptsize ±#1}}

\setlength{\aboverulesep}{0pt}
\setlength{\belowrulesep}{0pt}
\renewcommand{\arraystretch}{1.1}

\begin{tabular}{>{\centering}clccccc} 
\toprule
\rule[-0.55em]{0pt}{1.8em}Dataset & \rule[-0.55em]{0pt}{1.8em}Model & \rule[-0.55em]{0pt}{1.8em}PCC@10 $\uparrow$ & \rule[-0.55em]{0pt}{1.8em}PCC@50 $\uparrow$ & \rule[-0.55em]{0pt}{1.8em}PCC@200 $\uparrow$ & \rule[-0.55em]{0pt}{1.8em}MSE $\downarrow$ & \rule[-0.55em]{0pt}{1.8em}MAE $\downarrow$ \\
\midrule

\multirow{5}{*}{Kidney} 
& TRIPLEX & 0.579\std{0.095} & 0.485\std{0.084} & 0.351\std{0.066} & 1.122\std{0.204} & 0.855\std{0.104} \\
& StNet    & 0.523\std{0.105} & 0.435\std{0.095} & 0.305\std{0.064} & 1.167\std{0.217} & 0.847\std{0.078} \\
& BLEEP    & 0.518\std{0.112} & 0.434\std{0.102} & 0.310\std{0.071} & 1.233\std{0.244} & 0.865\std{0.085} \\
& Stem     & 0.535\std{0.111} & 0.414\std{0.084} & 0.271\std{0.059} & 1.380\std{0.347} & 0.911\std{0.115} \\
& \cellcolor{gray!20}\rule[-0.2em]{0pt}{1.1em}\textbf{HyperST}
 & \cellcolor{gray!20}\textbf{0.617\std{0.094}} 
 & \cellcolor{gray!20}\textbf{0.526\std{0.088}} 
 & \cellcolor{gray!20}\textbf{0.390\std{0.070}} 
 & \cellcolor{gray!20}\textbf{1.077\std{0.155}} 
 & \cellcolor{gray!20}\textbf{0.817\std{0.058}} \\
\midrule

\multirow{5}{*}{Colorectum} 
 & TRIPLEX& 0.701\std{0.128} & 0.624\std{0.154} & 0.462\std{0.191} & 1.869\std{0.803} & 1.056\std{0.239} \\
& StNet    & 0.646\std{0.134} & 0.570\std{0.142} & 0.419\std{0.176} & 1.686\std{0.373} & 1.023\std{0.134} \\
& BLEEP    & 0.637\std{0.112} & 0.556\std{0.120} & 0.382\std{0.160} & 2.038\std{0.587} & 1.096\std{0.164} \\
& Stem     & 0.670\std{0.116} & 0.573\std{0.130} & 0.399\std{0.166} & 1.788\std{0.418} & 1.032\std{0.138} \\
& \cellcolor{gray!20}\rule[-0.2em]{0pt}{1.1em}\textbf{HyperST}
 & \cellcolor{gray!20}\textbf{0.721\std{0.105}} 
 & \cellcolor{gray!20}\textbf{0.642\std{0.128}} 
 & \cellcolor{gray!20}\textbf{0.477\std{0.184}} 
 & \cellcolor{gray!20}\textbf{1.498\std{0.456}} 
 & \cellcolor{gray!20}\textbf{0.958\std{0.158}} \\
\midrule

\multirow{5}{*}{Skin} 
& TRIPLEX & 0.831\std{0.094} & 0.799\std{0.114} & 0.740\std{0.142} & 0.981\std{0.466} & 0.685\std{0.205} \\
& StNet    & 0.804\std{0.105} & 0.779\std{0.117} & 0.726\std{0.140} & 0.993\std{0.469} & 0.689\std{0.198} \\
& BLEEP    & 0.788\std{0.111} & 0.761\std{0.123} & 0.704\std{0.145} & 1.117\std{0.540} & 0.701\std{0.221} \\
& Stem     & 0.782\std{0.094} & 0.748\std{0.113} & 0.687\std{0.138} & 1.276\std{0.703} & 0.730\std{0.261} \\
& \cellcolor{gray!20}\rule[-0.2em]{0pt}{1.1em}\textbf{HyperST}
 & \cellcolor{gray!20}\textbf{0.839\std{0.086}} 
 & \cellcolor{gray!20}\textbf{0.812\std{0.102}} 
 & \cellcolor[HTML]{E0E0E0}\textbf{0.758\std{0.129}} 
 & \cellcolor{gray!20}\textbf{0.932\std{0.418}} 
 & \cellcolor{gray!20}\textbf{0.657\std{0.182}} \\
\midrule

\multirow{5}{*}{Lung}
& TRIPLEX & 0.567\std{0.247} & 0.499\std{0.272} & 0.393\std{0.272} & 1.537\std{1.307} & 0.849\std{0.446} \\
& StNet   & 0.526\std{0.247} & 0.464\std{0.267} & 0.355\std{0.253} & 1.660\std{1.258} & 0.899\std{0.422} \\
& BLEEP   & 0.488\std{0.212} & 0.412\std{0.235} & 0.311\std{0.230} & 1.803\std{1.507} & 0.891\std{0.465} \\
& Stem    & 0.546\std{0.224} & 0.469\std{0.250} & 0.351\std{0.241} & 1.709\std{1.365} & 0.866\std{0.485} \\
& \cellcolor{gray!20}\rule[-0.2em]{0pt}{1.1em}\textbf{HyperST}
 & \cellcolor{gray!20}\textbf{0.637\std{0.225}}
 & \cellcolor{gray!20}\textbf{0.568\std{0.260}}
 & \cellcolor{gray!20}\textbf{0.459\std{0.282}}
 & \cellcolor{gray!20}\textbf{1.182\std{0.873}}
 & \cellcolor{gray!20}\textbf{0.757\std{0.352}} \\
\bottomrule
\end{tabular}  
\end{table*}

\paragraph{Multi-Level Genomic Profiles Extractor} Let $Y_{s} \in \mathbb{R}^N$ be the associated spot-level gene expression profile of the spot-level image $X_{s}$, where $N$ is the gene set size. In the same vein as the niche-level image patch, we introduce the niche-level gene expression profile $ Y_{n} \in \mathbb{R}^N = \frac{1}{|S|} \sum_{i \in S}Y_{s}^i$, 
where $S=\{Y_{s}^1, \cdots, Y_{s}^K\}$ denotes the expression profile set of $Y_{s}$ and its neighbors, $K-1$ is the number of selected neighbors. $G_{s}, G_{n}=MGE(Y_{s}, Y_{n} ;\theta_{gene})$, 
where $MGE$ denotes the multi-level genomic profiles extractor with trainable parameters implemented by a trainable fully connected network $\theta_{gene}$, $G_{s} \in \mathbb{R}^{d}$ and $G_{n}\in \mathbb{R}^{d}$ are the spot-level and niche-level gene embeddings, respectively.

\subsection{Hierarchical Hyperbolic Alignment}
\label{Section 3.3}
To obtain better representations for facilitating the subsequent tasks, the alignment is a pivotal method  which bridges the gap of different modalities~\citep{xie2023spatially, zhang2025atomas, li2021align}. However, common implementations of alignment, such as BLEEP~\citep{xie2023spatially} , directly minimize distances between items in Euclidean space, which may not be appropriate for hierarchical data like ST.  To address this problem, we design a \textbf{Hierarchical Contrastive Alignment} module, which aligns the different modalities at different levels in hyperbolic space. Subsequently, we introduce a \textbf{Hierarchical Entailment Alignment} module to regularize the partial order in ST data.

\paragraph{Hierarchical Contrastive Alignment (HCA)} Using Equation \ref{exponential map}, let $\exp_O^c(\cdot):\mathbb{R}^d\to \mathbb{L}^d_c$ map Euclidean features to hyperbolic space with trainable curvature $-c<0$ and origin $O$. This yields hyperbolic spatial components $ \{\hat{I}_{s}^{space},\hat{I}_{n}^{space}, \hat{G}_{s}^{space},\hat{G}_{n}^{space} \}=\exp_{O}^{c}{(\{I_{s}, I_{n}, G_{s}, G_{n}\})}$,
while the corresponding time components can be calculated by Equation \ref{hyperbolic denfine}. The hyperbolic representations $\hat{I}_{s}$, $\hat{I}_{n}$, $\hat{G}_{s}$ and $\hat{G}_{n}$ are obtained by concatenating spatial components and time components. To align the spot-level image embedding to the spot-level  embedding, we employ a modified infoNCE loss~\citep{oord2018representation}, in which the cosine similarity is replaced by the Lorentzian distance $d_{\mathbb{L}} (\cdot, \cdot)$ described in Equation \ref{distence of hyperbolic}. The contrastive loss is defined as follows:
\begin{equation}
    \mathcal{L}_{align}(\hat{I}_{s}, \hat{G}_{s}) = - \frac{1}{B}\sum_{i=1}^{B}\log\frac{\exp(d_{\mathbb{L}}(\hat{I}_{s}^i,\hat{G}_{s}^{i})/\tau)}{\sum_{j=1,j\neq i}^B\exp(d_{\mathbb{L}}(\hat{I}_{s}^i,   \hat{G}_{s}^{j})/\tau)},
\end{equation}
where $B$ denotes the batch size and $\tau$ is the temperature parameter. To better utilize in-batch negatives, we also align spot-level gene to spot-level image embeddings using $\mathcal{L}_{align}(\hat{G}_{s}, \hat{I}_{s})$. Since spot-level features represent more general characteristics, a single spot-level feature may correspond to multiple niche-level features within a batch. To avoid such undesirable negative alignment, we only consider the alignment from niche-level features to spot-level features, i.e., $\mathcal{L}_{align}(\hat{G}_{n}, \hat{I}_{s})$ and $\mathcal{L}_{align}(\hat{I}_{n}, \hat{G}_{s})$. The objective function of Hierarchical Contrastive Alignment can be expressed as:

\begin{equation}
  \label{eq:HCA_loss}
  \begin{aligned}
    \mathcal{L}_{HCA} = \frac{1}{4} ( & \mathcal{L}_{align}(\hat{I}_{s}, \hat{G}_{s}) + \mathcal{L}_{align}(\hat{G}_{s}, \hat{I}_{s}) \\
                                    & + \mathcal{L}_{align}(\hat{G}_{n}, \hat{I}_{s}) + \mathcal{L}_{align}(\hat{I}_{n}, \hat{G}_{s}) ) .
  \end{aligned}
\end{equation}

\paragraph{Hierarchical Entailment Alignment (HEA)} Beyond spot-niche hierarchies, we account for the non-identical nature of image features and gene features. We recognize that gene features provide finer-grained molecular insights. Thus, we posit that gene features are the child concept of images in hyperbolic space. In our ST data, this hierarchy can be summarized  as spot-level features entailing niche-level features, and pathological images entailing their corresponding gene expression profiles. In order to directly constrain this hierarchical structure, we leverage Hyperbolic Entailment Loss $\mathcal{L}_{entail}(\cdot,\cdot)$ described in Equation \ref{entail loss}. Therefore, the final objective function of this module is formulated as:

\vspace{-3.5mm}

\begin{equation}
  \label{eq:HEA_loss}
  \begin{aligned}
    \mathcal{L}_{HEA} = \frac{1}{4} ( & \mathcal{L}_{entail}(\hat{I}_{s}, \hat{I}_{n}) + \mathcal{L}_{entail}(\hat{G}_{s}, \hat{G}_{n}) \\
                                    & + \mathcal{L}_{entail}(\hat{I}_{s}, \hat{G}_{s}) + \mathcal{L}_{entail}(\hat{I}_{n}, \hat{G}_{n}) ) .
  \end{aligned}
\end{equation}

\subsection{Gene Decoder Based on Aligned Representations and Objective Function}
\label{Section 3.4}
To predict the spot-level gene expression profiles, we directly concatenate the aligned representations ( $I_s$ and $I_n$) and feed the result into a gene decoder implemented by Multi-Layer Perceptron (MLP), which can be expressed by $Y^{pred}=Decoder_{gene}(\text{concat}(I_s,I_n))$. MSE loss is leveraged to optimize this decoder: $\mathcal{L}_{pred}=\| Y^{pred} - Y_s\|_2^2$.

\paragraph{The Training Objective Function} 

The training objective of HyperST is twofold: (1) to align pathological images and gene expression profiles across multiple levels by modeling the hierarchical structure of ST data, and (2) to accurately predict gene expression from image features alone. 
%
This objective function consists of two components: hierarchical alignment loss and ST prediction loss, defined by:
\begin{equation}
    \mathcal{L}=\mathcal{L}_{pred} + \alpha (\mathcal{L}_{HCA}+\beta \mathcal{L}_{HEA}),
\end{equation}
where $\alpha$ balances the loss components, and $\beta$ controls the entailment loss effect.

\section{Experiments and Results}\label{exper}

\begin{figure*}[t] 
    \centering
    \includegraphics[width=1\textwidth]{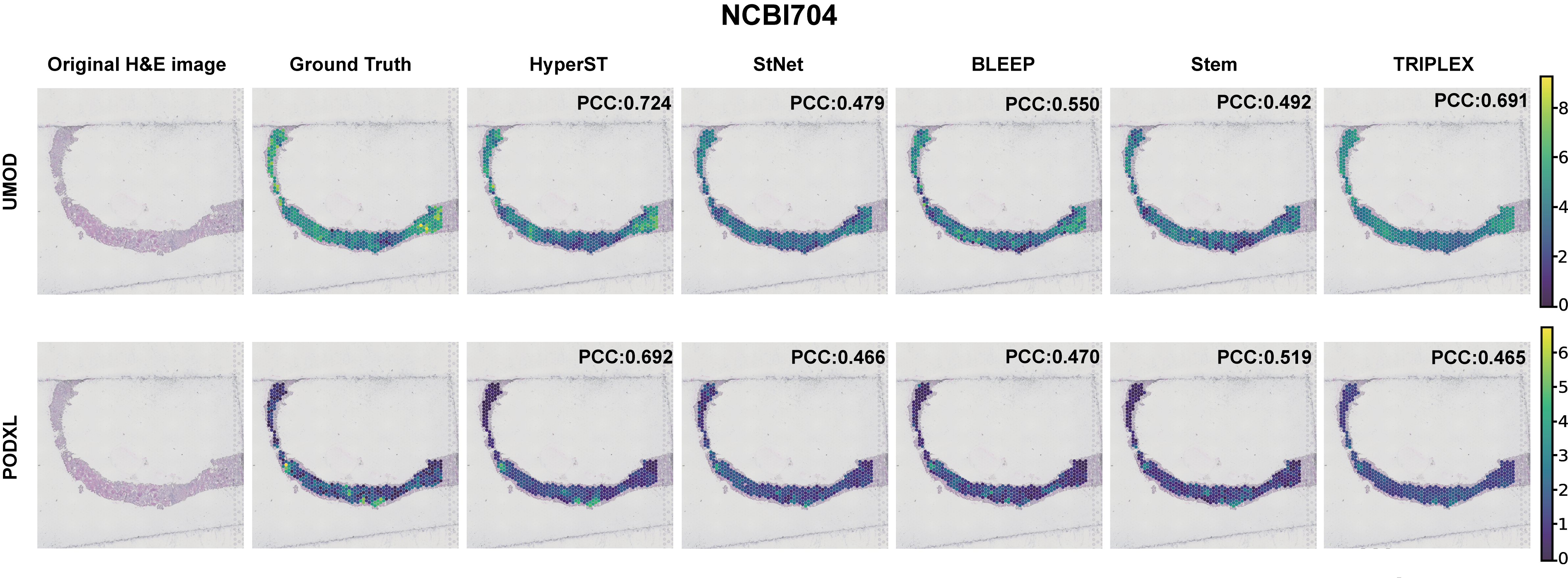}  
    \caption{Visualization of the spatial distribution of the UMOD (Top) gene and PODXL gene (Bottom) in NCBI704 sample. The color scale ranges from purple/blue (low expression) to yellow/green (high expression).}
    \label{VIS1}
\end{figure*}

\subsection{Experimental Settings}
\label{exper_1}

\paragraph{Dataset} 

To evaluate HyperST, we collected four public datasets from the HEST-1K dataset~\citep{jaume2024hest}, a high-quality collection of spatial transcriptomics data with standardized processing and rich metadata. (1) Kidney~\citep{lake2023atlas} provides 23 WSIs and 25,944 spots at a resolution of approximately 0.76 $\mu$m/pixel. (2) Colorectum dataset~\citep{valdeolivas2024profiling} comprises 14 WSIs (0.45 $\mu$m per pixel) with a total of 20,733 spots. (3) Skin~\citep{schabitz2022spatial} includes 46 WSIs and over 35,000 spots. The resolution of pathology images is about 0.52 $\mu$m/pixel. (4) Lung ~\citep{madissoon2023atlas}
consists of 16 WSIs at 0.45 $\mu$m/pixel resolution and 31,445 spots. The spots of all datasets have a 
diameter of 55 $\mu$m.

\paragraph{ST Preprocessing}
To account for variations in image resolution across datasets, we adopted a physics-aware patch extraction strategy rather than using fixed pixel dimensions for image cropping. 
Specifically, we calculated the patch size for each spot based on its physical diameter and cropped the corresponding images at their respective resolutions to obtain spot-level image patches. 
%
The niche-level patch is created by cropping the region encompassing the central spot and its K-nearest neighbors (determined by spatial coordinates). 
Subsequently, all extracted patches are resized to a uniform 224×224 pixel resolution. For the gene expression data, we select the top 200 Highly Mean, Highly Variant Genes (HMHVG). Gene expression counts for each spot were subsequently log-transformed.

\paragraph{Evaluation Protocol}
\label{Evaluation Protocol}
To ensure robust model evaluation, we performed five independent random splits of the WSI samples for each dataset, allocating 80\% for training, 10\% for validation, and 10\% for testing in each iteration. The exact WSI IDs used for each of the five splits are provided in our code to ensure full reproducibility. Our evaluation metrics include top-$k$ mean Pearson Correlation Coefficient (PCC@$k$), mean squared error (MSE), and mean absolute error (MAE), similar to~\citep{zhu2025diffusion, chung2024accurate}.

\subsection{Experimental Results and Visualization}
\paragraph{Baseline Comparison}

Table~\ref{tab:baseline} shows that HyperST outperforms all existing methods across all datasets (Kidney, Colorectum, Skin and Lung), highlighting HyperST’s superior accuracy and robustness in predicting gene expression from pathology images across diverse biological contexts, which were trained under a rigorous, fair comparison protocol (details in the Section 7.2 of Supplementary Material). 
To emphasize the magnitude of this improvement, we compare HyperST to the second-best performing model, TRIPLEX. On the PCC@200 metric, HyperST demonstrates substantial relative improvements of approximately 10.95$\%$, 3.24$\%$, 2.52$\%$, and 16.7$\%$ on the Kidney, Colorectum, Skin, and Lung datasets, respectively.
Notably, while the next-best TRIPLEX confirms the value of multi-level features, it lacks explicit constraints for the intrinsic hierarchy. HyperST’s lead demonstrates the crucial advantage of our Hierarchical Hyperbolic Alignment.


\paragraph{Biomarker Visualization}

To further qualitatively assess the model's behavior, we performed visualization of sample  NCBI704 in the Kidney dataset, focusing on two established kidney cancer biomarkers, i.e., UMOD~\citep{turner2021umod} and PODXL~\citep{barua2014exome}. Figure \ref{VIS1} demonstrates that HyperST more accurately captures the key high-expression regions.

\begin{table}[htbp]
    \centering
    \caption{Performance on MSI Status Classification (AUROC). Higher values on MSI-H and MSS are better.}
    \label{tab:msi_status_classification_wrap}
    \setlength{\aboverulesep}{0pt}
    \setlength{\belowrulesep}{0pt}
    \renewcommand{\arraystretch}{1.1}

    \begin{tabular}{lcc}
        \toprule
        \multirow{2}{*}{\rule[-0.55em]{0pt}{1.8em}Model} & \multicolumn{2}{c}{AUROC} \\
        \cmidrule(lr){2-3}
        & \rule[-0.55em]{0pt}{1.8em}MSI-H $\uparrow$ & MSS $\uparrow$ \\
        \midrule
        TRIPLEX & 0.630{\scriptsize ±0.048} & 0.567{\scriptsize ±0.032} \\
        StNet   & 0.571{\scriptsize ±0.080} & 0.541{\scriptsize ±0.029} \\
        BLEEP   & 0.550{\scriptsize ±0.045} & 0.530{\scriptsize ±0.056} \\
        Stem    & 0.584{\scriptsize ±0.078} & 0.544{\scriptsize ±0.034} \\
        \rowcolor{gray!20}
        \textbf{HyperST} 
        & \textbf{0.719{\scriptsize ±0.060}} 
        & \textbf{0.601{\scriptsize ±0.056}}\\
        \bottomrule
    \end{tabular}
\end{table}

\begin{figure*}[htbp]
    \centering

    \includegraphics[width=1\textwidth]{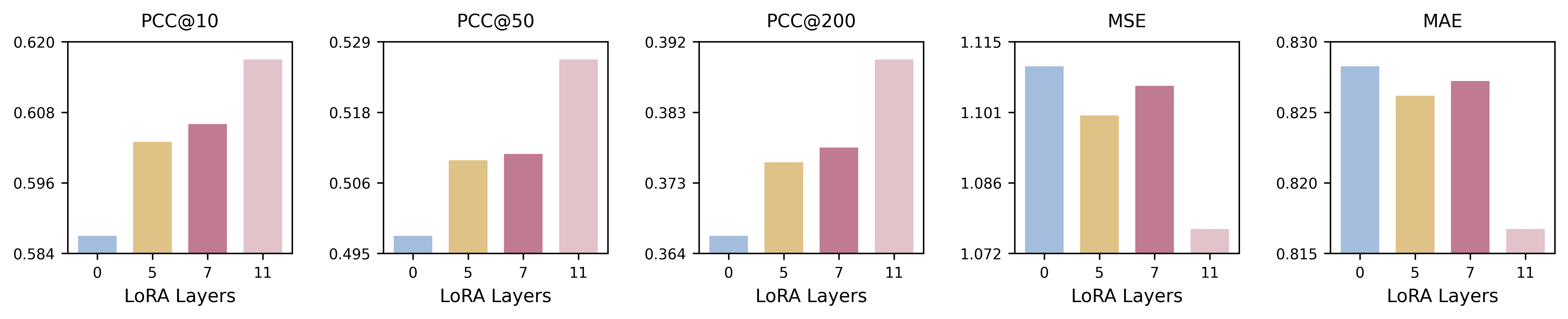}
    
    \caption{Result of ablation study on the choice of the last layers of LoRA.}
    \label{fig:lora_subplot}

\end{figure*}

\begin{table*}[t]
\centering
\caption{Ablation study of the alignment strategy.}
\label{Table 3: ablation of alignment}
\setlength{\aboverulesep}{0pt}
\setlength{\belowrulesep}{0pt}
\renewcommand{\arraystretch}{1.1}

\begin{tabular}{lccccc}
\toprule
\rule[-0.55em]{0pt}{1.8em}Alignment & \rule[-0.55em]{0pt}{1.8em}PCC@10 $\uparrow$ & \rule[-0.55em]{0pt}{1.8em}PCC@50 $\uparrow$ & \rule[-0.55em]{0pt}{1.8em}PCC@200 $\uparrow$ & \rule[-0.55em]{0pt}{1.8em}MSE $\downarrow$ & \rule[-0.55em]{0pt}{1.8em}MAE $\downarrow$ \\
\midrule

w/o G-I HEA    & 0.610{\scriptsize ±0.101} & 0.514{\scriptsize ±0.096} & 0.378{\scriptsize ±0.074} & 1.147{\scriptsize ±0.188} & 0.839{\scriptsize ±0.065} \\
w/o HEA        & 0.603{\scriptsize ±0.092} & 0.508{\scriptsize ±0.083} & 0.368{\scriptsize ±0.061} & 1.112{\scriptsize ±0.185} & 0.831{\scriptsize ±0.065} \\
w/o HEA + HCA  & 0.576{\scriptsize ±0.099} & 0.484{\scriptsize ±0.089} & 0.344{\scriptsize ±0.064} & 1.134{\scriptsize ±0.168} & 0.837{\scriptsize ±0.058} \\
\arrayrulecolor{gray!10}\noalign{\hrule height 0.2pt}
\arrayrulecolor{black}

MERU           & 0.586{\scriptsize ±0.099} & 0.494{\scriptsize ±0.090} & 0.355{\scriptsize ±0.062} & 1.148{\scriptsize ±0.205} & 0.842{\scriptsize ±0.071} \\
CLIP           & 0.558{\scriptsize ±0.098} & 0.462{\scriptsize ±0.087} & 0.321{\scriptsize ±0.058} & 1.220{\scriptsize ±0.293} & 0.867{\scriptsize ±0.097} \\

\arrayrulecolor{gray!10}\noalign{\hrule height 0.2pt}
\arrayrulecolor{black}

\cellcolor{gray!20}\rule[-0.2em]{0pt}{1.1em}\textbf{Ours}
 & \cellcolor{gray!20}\textbf{0.617{\scriptsize ±0.094}}
 & \cellcolor{gray!20}\textbf{0.526{\scriptsize ±0.088}}
 & \cellcolor{gray!20}\textbf{0.390{\scriptsize ±0.070}}
 & \cellcolor{gray!20}\textbf{1.077{\scriptsize ±0.155}}
 & \cellcolor{gray!20}\textbf{0.817{\scriptsize ±0.058}} \\
\arrayrulecolor{black}\bottomrule
\end{tabular}
\end{table*}

\vspace{-3mm}

\paragraph{Clinical Downstream Task Validation}

We designed a downstream validation experiment to further validate the clinical utility of HyperST’s zero-shot gene expression predictions compared with other methods. 
First, we employed our model, pre-trained on the Colorectum dataset, to perform zero-shot inference on H\&E slides from an independent external dataset, TCGA-COADREAD (colon and rectal adenocarcinoma). 
For each slide, HyperST predicts a gene expression vector for every spot. We then obtain a slide-level pseudo-bulk profile by averaging the predicted expression across all spots on that slide.
The resulting pseudo-bulk profiles were then used to train a Random Forest classifier for predicting microsatellite instability (MSI) status, a critical clinical biomarker for immunotherapy response~\citep{feng2024spatially}. 
As shown in Table~\ref{tab:msi_status_classification_wrap}, the gene profiles predicted by HyperST led to significantly better MSI prediction performance compared to other baselines by capturing more clinically relevant signals. 
In this zero-shot setting, HyperST attains per-class AUROCs of 0.719 for MSI-H and 0.601 for MSS. 
Compared with the strongest baseline TRIPLEX (0.630 and 0.567, respectively), HyperST achieves approximately 14$\%$ and 6$\%$ higher AUROC
for MSI-H and MSS.

\subsection{Ablation Study}


We performed an ablation study on the model's structure and hyperparameters to observe the strategy of alignment, input data for gene decoder and the impact of LoRA. Here, we describe the results on Kidney dataset. 


\vspace{-2.5mm}

\paragraph{Strategy of Alignment} 
We compared the different alignment strategies including: a) removing only the gene-image regularization term of the HEA loss (w/o G-I HEA), b) removing the entire HEA loss (w/o HEA), c) removing the entire Hierarchical Hyperbolic Alignment (HHA) module (w/o HEA + HCA), d) replacing the HHA module by a MERU variant in Hyperbolic Space (without mutil-level representation learning)~\citep{desai2023hyperbolic} and e) replacing the HHA module by a CLIP variant in Euclidean Space~\citep{radford2021learning}.
As shown in Table \ref{Table 3: ablation of alignment}, the results confirm our design. Removing the entire Hierarchical Hyperbolic Alignment module caused a 13.26$\%$ PCC@200 drop. Removing the entire HEA loss or just its gene-image regularization term led to 6.01$\%$ and 3.24$\%$ drops. Notably, the performance gap between our full model and its Euclidean counterpart (CLIP) strongly validates our core hypothesis on the superiority of hyperbolic space for this task.

\vspace{-1.5mm}

\paragraph{Input of Decoder} We evaluated the impact of different input strategies on the decoder's performance, including using a) only spot-level image, b) only niche-level image and c) the combination of both. As shown in Table \ref{tab:decoder_input_ablation_kidney}, the results reveal that our combined approach yields the best performance across all metrics.

\begin{table}[htbp]
\centering
\setlength{\tabcolsep}{5pt}
\caption{Additional ablation study of the decoder input.}
\label{tab:decoder_input_ablation_kidney}
\setlength{\aboverulesep}{0pt}
\setlength{\belowrulesep}{0pt}
\renewcommand{\arraystretch}{1.1}

\begin{tabular}{lccc}
\toprule
\rule[-0.55em]{0pt}{1.8em}Decoder Input & \rule[-0.55em]{0pt}{1.8em}PCC@200 $\uparrow$ & \rule[-0.55em]{0pt}{1.8em}MSE $\downarrow$ & \rule[-0.55em]{0pt}{1.8em}MAE $\downarrow$ \\
\midrule

only spot     & 0.353{\scriptsize ±0.067} & 1.171{\scriptsize ±0.184} & 0.849{\scriptsize ±0.065} \\
only niche    & 0.356{\scriptsize ±0.073} & 1.110{\scriptsize ±0.139} & 0.828{\scriptsize ±0.051} \\
\rowcolor{gray!20}
\textbf{spot+niche} 
& \textbf{0.390{\scriptsize ±0.070}} 
& \textbf{1.077{\scriptsize ±0.155}} 
& \textbf{0.817{\scriptsize ±0.058}} \\
\bottomrule
\end{tabular}

\end{table}

\paragraph{Choice of LoRA} During training, we found that increasing the number of the last attention layers in UNI adapted by LoRA modules improved performance. The corresponding evaluation results are shown in Figure \ref{fig:lora_subplot}. Our ablation study demonstrates that increasing LoRA layers from 0 to 11 layers generally improves predictive performance across most metrics. However, we observe marginal degradation in MSE and MAE when transitioning from 5 to 7 layers.

\section{Conclusion}
We present HyperST, a novel framework that leverages  multi-level hyperbolic representations to predict spatial transcriptomics from histology images. By modeling the intrinsic hierarchical structure of ST data within hyperbolic space, HyperST learns more comprehensive spatial histological and genetic features.
Our comprehensive experimental evaluation demonstrates that HyperST consistently outperforms state-of-the-art approaches, underscoring the potential of geometric deep learning in spatial omics analysis.
%


\clearpage
{
    \small
    \bibliographystyle{ieeenat_fullname}
    \bibliography{main}
}




\end{document}